%% file: fale_ue23.tex
\documentclass[runningheads]{llncs}
\usepackage[T1]{fontenc}
\usepackage{graphicx}
\usepackage{cite}

\usepackage{amssymb}

\usepackage{bm}

\def\vx{{\bm{x}}}
\def\vz{{\bm{z}}}
\def\X{{\bm{\mathcal{X}}}}

\def\D{{\mathcal{D}}}

\def\R{\mathbb{R}}
\def\E{\mathbb{E}}

\def\ale{\mathsf{ALE}}
\def\fale{\mathsf{FALE}}

\newcommand\blfootnote[1]{%
  \begingroup
  \renewcommand\thefootnote{}\footnote{#1}%
  \addtocounter{footnote}{-1}%
  \endgroup
}

\begin{document}
\title{FALE: Fairness-Aware ALE Plots for Auditing Bias in Subgroups}

\author{Giorgos Giannopoulos\inst{1}\orcidID{0000-1111-2222-3333}
%\orcidID{0000-1111-2222-3333} 
\and
Dimitris Sacharidis\inst{2} \and
Nikolas Theologitis\inst{1} \and
Loukas Kavouras\inst{1} \and
Ioannis Emiris\inst{1}}
\authorrunning{G. Giannopoulos et al.}
\institute{Athena Research Center, Greece  \and
Université Libre de Bruxelles, Belgium
}

\maketitle              % typeset the header of the contribution
\begin{abstract}
Fairness is steadily becoming a crucial requirement of Machine Learning (ML) systems. A particularly important notion is \textit{subgroup fairness}, i.e., fairness in subgroups of individuals that are defined by more than one attributes. Identifying bias in subgroups can become both computationally challenging, as well as problematic with respect to comprehensibility and intuitiveness of the finding to end users. In this work we focus on the latter aspects; we propose an explainability method tailored to identifying potential bias in subgroups and visualizing the findings in a user friendly 
%and interactive 
manner to end users. In particular, we extend the ALE plots explainability method, proposing FALE (Fairness aware Accumulated Local Effects) plots, a method for measuring the change in fairness for an affected population corresponding to different values of a feature (attribute). We envision FALE to function as an efficient, user friendly, comprehensible and reliable first-stage tool for identifying subgroups with potential bias issues.
\blfootnote{Preprint. Presented in Uncertainty meets Explainability Workshop @ ECML/PKDD 2023.} 

\keywords{AI fairness  \and Explainability \and Subgroup fairness.}
\end{abstract}

\input{intro}

\input{related-work}

\input{method}

\input{results}

\bibliographystyle{splncs04}
\bibliography{bibliography}

\end{document}

%% file: intro.tex
\section{Introduction}
\label{sec:intro}

Fairness in AI is an open problem that has attracted the attention of the research community the last years. Although there exist a plethora of methods for auditing bias in AI, there does not exist a one-size-fits-all solution to the problem \cite{WACHTER2021105567, HAUER2021105583}. In particular, various statistical definitions have been proposed \cite{10.1145/3194770.3194776} that examine the statistics of an AI system's prediction (considering or not the statistics of the ground truth) with respect to a specific sensitive attribute (sex, gender, race, age, etc). A standard practice is to select a sensitive attribute and a fairness definition and evaluate the unfairness of the model on a given test set.

Nevertheless it is often the case that bias can be identified in a minority group defined by more than one attributes, sensitive or note. This is the case of \textit{subgroup} (alt.\ intersectional, multidimensional) fairness \cite{pmlr-v80-kearns18a, WACHTER2021105567}. Consider for example a scenario where an AI 
system decides whether a person is promoted based on a specific feature set and we want to audit the fairness of the system with respect to two sensitive attributes: \textit{gender}, with values $\{$\textit{male}, \textit{female}$\}$ and \textit{race}, with values $\{$\textit{caucasian}, \textit{non-caucasian}$\}$. 
%We select the statistical definition of 
%\textit{equalized odds}, that demands, for each sensitive attribute, that the protected and unprotected groups have equal \textit{true positive rates} and equal
%\textit{false positive rates} respectively. 
In our scenario, the unprotected groups are defined by \textit{female} and \textit{non-caucasian} respectively. It 
might be the case that auditing fairness individually on the two sensitive attributes finds the promotion decisions of the system fair, however, by further 
examining the result, one might identify that non-caucasian males and caucasian females are disproportionally unfavored compared to the other two subgroups, i.e., 
caucasian males and non-caucasian females.

In the above toy example, it is straightforward and computationally feasible to exhaustively investigate the subgroups defined by the two attributes. However, in real world settings which involve much larger feature sets, as well as feature value ranges, this task becomes infeasible. Further there is always the risk of gerrymandering, which consists in partitioning attributes in such way and granularity to mask bias in subgroups. These issues are examined by certain works in the literature, with the most prominent being \cite{pmlr-v80-kearns18a}. There, the authors provide an efficient method for auditing arbitrary subgroups on fairness, as well as a method for learning respectively fair classifiers. 
Other works also touch the concept of subgroup fairness. Indicatively, 
\cite{Tramr2015DiscoveringUA} present a framework for debugging associations between protected attributes and model output, that imply bias, in subgroups. \cite{10.1145/3351095.3372845} employ optimal transport to rank attributes that are mostly associated with different model behaviors in different protected subgroups.

A common issue with all aforementioned works is that they hardly focus on presenting their findings in a user intuitive and explainable way. In this paper, we propose a method that addresses this issue. We build on a visual explainability method, ALE plots, that are used to visualize the \textit{correlation-free} influence (i.e., the local effect) of model features on the average prediction of the model. We extend them in order to calculate, instead of the average model prediction, the measured fairness with respect to a selected statistical fairness definition \cite{10.1145/3194770.3194776} and a sensitive attribute. This quantity is used to measure the influence of another selected attribute (\textit{examined attribute}) on fairness. The changes of the measured fairness for different subgroups defined with respect to the examined attribute are visualized in a plot, along with the populations of the individuals within these subgroups. Our method can serve as a first-step auditing tool to help users quickly point their focus on specific subgroups to investigate fairness issues.

%With the realization that AI systems are able to pick existing bias in the training data and discriminate against various protected groups and subgroups, 
%defined by sex, sexual orientation, race, ethnic origin, religion, age, disabilities, etc. 
%research community is increasingly focusing on methods for defining, identifying, measuring and correcting bias in data and in algorithms.
%-correlation free
%-agility to vary bins and thus avoid gerrymandering
%-2D can visualize the joint contribution of 2 features on unfairness

%% file: related-work.tex
\section{Related Work}
\label{sec:related-work}

There is a line of work on auditing models for fairness of predictions at the subpopulation level \cite{kearns2018preventing, kearns2019empirical}. For example, \cite{tramer2017fairtest} identifies subpopulations that show dependence between a performance measure and the protected attribute. \cite{black2020fliptest} determines whether people are harmed due to their membership in a specific group by examining a ranking of features that are most associated with the model's behavior. 
There is no equivalent work for fairness of recourse, although the need to consider the subpopulation is recognized in \cite{karimi2020algorithmic} due to uncertainty in assumptions or to intentionally study fairness.

ALE plots are taxonomized as global explainability methods, since they explain the predicted outcome of the model on the whole population. The type of explanation is visual by drawing plots that showcase the influence of attributes to the predicted outcome. Other global explainability methods that use plots as explanations are the PDP plots and the M-plots \cite{friedman2000greedy, greenwell2018simple}. Other methods that explain model predictions by reporting the importance of features such as the Shapley values \cite{pmlr-v119-sundararajan20b} can also be directly extended to audit subgroup fairness by averaging the importance of all features for each individual and by examining if sensitive attributes play a significant role in model decisions. 

Another line of work that is related to ours in terms of subgroup fairness as well as global explainability makes use of counterfactual explanations. For example, recourse summaries \cite{rawal2020beyond,ley2022global} summarizes individual counterfactual explanations \emph{globally}, and as the authors in \cite{rawal2020beyond} suggest can be used to manually audit for unfairness in subgroups of interest. \cite{lakkaraju2019faithful} aims to explain how a model behaves in subspaces characterized by certain features of interest. \cite{cornacchia2023auditing} uses counterfactuals to unveil whether a black-box model, that already complies with the regulations that demand the omission of sensitive attributes, is still biased or not, by trying to find a relation between proxy features and bias.

%% file: method.tex
\section{Fairness-Aware ALE Plots}

\paragraph{Preliminaries.}
ALE plots compute and visualize the local effect of a feature to the output of an ML model.
We consider a \emph{feature space} $\X = X_1 \times \dots \times X_n$ of $n$ features. For an instance $\vx \in X$, we use the notation $\vx.X_i$ to refer to its value in feature $X_i$. We assume a \emph{model} $f$ maps an input $\vx \in \X$ to an output $f(\vx) \in \R$. 

The \emph{local effect} of feature value $x_i \in X_i$ for some instance $\vx$ such that $\vx.X_i=x_i$ is defined as the partial derivative of the model's output with respect to feature $X_i$, computed at $\vx$, i.e., $\frac{\partial f(\vx)}{\partial x_i}$. Then, the \emph{mean local effect} of feature value $x_i \in X_i$ is computed by averaging local effects for all instance that take value $x_i$ in their $X_i$ feature, i.e., $\E\left[\frac{\partial f(\vx)}{\partial x_i} \ | \ \vx.X_i = x_i \right]$. The \emph{accumulated local effect} (ALE) of feature value $x_i \in X_i$ is computed by summing up the local effects at all feature values of $X_i$ up to $x_i$: $\ale(x_i) = \int_{X_i.{min}}^{x_i} \E\left[\frac{\partial f(\vz)}{\partial z_i} \ | \ \vz.X_i = z_i \right]dz_i$, where $X_i.min$ refers to a value smaller than the minimum value in feature $X_i$.

In practice, the ALE values are estimated by the following procedure assuming a dataset $\D$ of instances. The domain of the feature $X_i$ is partitioned into $n$ bins, $\{x_i^0, \dots, x_i^n\}$, where the $k$-th bin is $b_i^k = (x_i^{k-1}, x_i^k]$. The local effect of value $x_i \in b_i^k$ that falls in the $k$-th bin for some instance $\vx$ is estimated as the difference of the model's output between instances $\vx^h$ and $\vx^l$, which take the same values as $\vx$ in all features except $X_i$, where they take the bin boundary values, i.e., $\vx^h.X_i = x_i^k$, and $\vx^l.X_i = x_i^{k-1}$. The mean local effect is estimated using the dataset instances that fall in each bin, and the ALE value of $x_i$ sums up these estimates over all bins up to the one, say $k(x_i)$, that covers $x_i$, i.e., $\displaystyle \hat\ale(x_i) = \sum_{k=1}^{k(x_i)} \frac{1}{|b_i^k|} \sum_{\vx\in \D : \vx.X_i \in b_i^k } \left( f(\vx^h) - f(\vx^l) \right) $, where $|b_i^k|$ denotes the number of dataset instances whose $X_i$ value falls in bin $b_i^k$. The \emph{ALE plot} for feature $X_i$ draws these ALE estimates as a function of feature values, after normalizing the estimates to have mean zero.

\paragraph{Fairness.}
Let $A$ denote the protected attribute, which, for ease of presentation, takes two values $\{0, 1\}$, where $1$ signifies the protected group. 
Consider an \emph{unfairness measure} $u(G^0, G^1)$ that quantifies the disparity of outcomes between a non-protected $G^0$ and a protected group $G^1$ according to some notion of fairness. As an example consider statistical parity, and assume a binary classifier $f$ where output 1 is the desired outcome. Then, an unfairness measure for statistical parity is $u(G^0, G^1) = \left| \frac{1}{|G^0|}\sum_{\vx \in G^0} f(\vx) - \frac{1}{|G^1|}\sum_{\vx \in G^1} f(\vx) \right|$.

\paragraph{Fairness-Aware ALE.}
Our goal is to study the effect of particular feature value $x_i \in X_i$ on unfairness.
For this purpose, let $G^0_{x_i} = \{ \vx \in D \ |\ \vx.X_i = x_i \wedge d \vx.A = 0 \}$ denote the non-protected group of instances that take value $x_i$ in the $X_i$ feature; the protected group $G^1_{x_i}$ is defined analogously.
More precisely, our goal is to quantify the effect of $x_i$ on the unfairness measure $u(G^0_{x_i}, G^1_{x_i})$, using the ALE framework.

As in ALE, we assume a bin-partitioning on feature $X_i$. 
Define two hypothetical groups $G^0_{k,l}$, $G^0_{k,h}$ for the non-protected instances that fall in the $k$-th bin; the former replaces the $X_i$ value of all instances with the lower bin boundary $x_i^{k-1}$, and the latter with the higher bin boundary $x_i^k$. Similarly, define their counterparts $G^1_{k,l}$, $G^1_{k,h}$ for the protected instances.
Then, we define the FALE estimate at feature value $x_i$ as:
\[
\hat\fale(x_i) = \sum_{k=1}^{k(x_i)}  \left( u(G^0_{k,h}, G^1_{k,h}) - u(G^0_{k,l}, G^1_{k,l}) \right).
\]
The \emph{FALE plot} for feature $X_i$ draws these FALE estimates as a function of feature values, after normalizing the estimates to have mean zero.

%% file: results.tex
\section{Demonstration}
\label{sec:demo}

\begin{figure}[h]
    \centering
    \begin{tabular}{cc}
        %\textbf{\scriptsize \quad \quad Ground Truth  \quad Model Prediction}\textbf{\scriptsize \quad \quad Ground Truth \quad Model Prediction} \\
        \includegraphics[width=\textwidth]{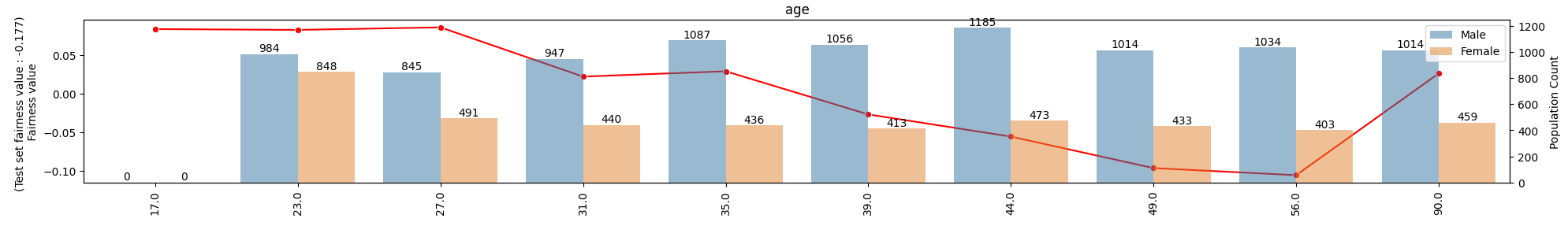} \\
         \textbf{\scriptsize Fig. 1 (a) FALE plot for the contribution of age}\\
        \includegraphics[width=\textwidth]{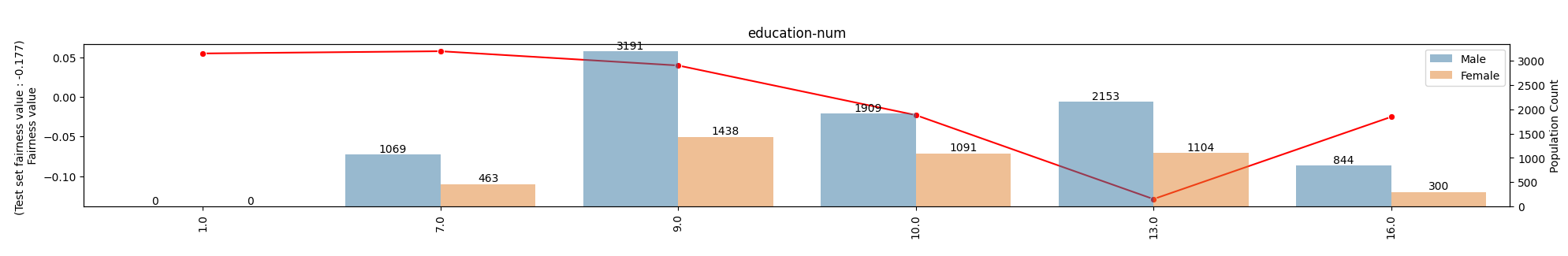} \\
         \textbf{\scriptsize Fig. 1 (b) FALE plot for the contribution of education}\\
        \includegraphics[width=\textwidth]{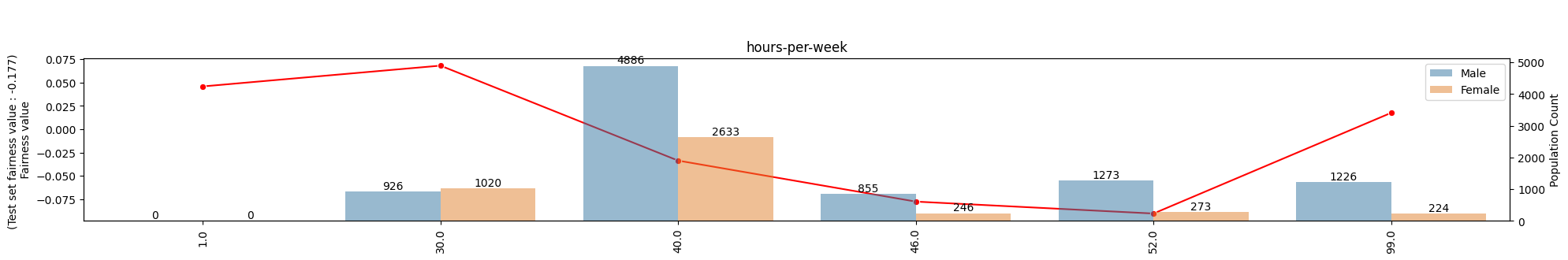} \\
         \textbf{\scriptsize Fig. 1 (c) FALE plot for the contribution of hours per week}\\
    \end{tabular}
    
    \caption[short]{FALE plots for the fairness definition of \textit{statistical parity} on the sensitive attribute sex in the adult dataset. Each FALE plot  examines the influence of a different attribute. Fig. 1 (a) depicts the contribution of the \textit{age} attribute, Fig. 1 (b), depicts the contribution of the \textit{education} attribute and, Fig. 1 (c) illustrates the influnce of the \textit{hours per week} attribute. Positive FALE effects indicate additional bias from the model towards females,  whereas negative FALE effects indicate less bias towards females compared to the average population. }
\label{fig:fale_sex_sp}
\end{figure}

In this section, we present some indicative plots that are produced by FALE. We 
use the Adult dataset \cite{Bache+Lichman:2013}. The experimental setting is as follows. We perform a standard train-test split and train and evaluate an XGBoost classifier, on the binary classification problem of predicting whether an instance's outcome is either less than or greater than 50K. We select a fairness definition and a sensitive attribute to examine on the test set. In the example of Figure \ref{fig:fale_sex_sp}, we choose the \textit{statistical parity} fairness definition on the \textit{sex} sensitive attribute. As examined attributes, we choose \textit{age}  \textit{education level} and \textit{hours per week} as shown respectively in Fig~\ref{fig:fale_sex_sp}.

The $y$-axis in each plot represents the increase or decrease of the measured fairness in each subgroup of the examined attribute, as compared to the measured fairness value in the whole test set. The parenthesis in the $y$-axis legend provides the latter value. In our example, the model predictions on the test set are biased against Female, by a value of $0.177$, according to statistical parity. Within each plot, following the edges of the red plot, we can see whether the measured unfairness is further increased, or reduced. For example, in Figure \ref{fig:fale_sex_sp} (a), we can see that unfairness towards females is significantly increased for middle-aged subgroups and slightly decreased for young-aged subgroups compared to the "average" population of the whole test.  The pairs of bars at each subgroup provide population information for the individual subgroups, to provide an extra intuition about the significance of the observed difference in fairness.

\section{Conclusion}
\label{sec:concl}
In this paper we presented FALE, a novel method for fast and user intuitive first-step auditing of fairness of ML models in subgroups. Our future work includes implementing 2D FALE, which will allow auditing more fine-grained subgroups, as well as further delve into and compare the semantics of additional visual explainability methods (PDP, M-plots) \cite{unknown}, in the context of fairness auditing. 

\subsubsection{Acknowledgements}
This work has been funded by the European Union’s Horizon Europe research and innovation programme under Grant Agreement No. 101070568 (AutoFair).

%The answer depends on the specific use cases such language models are deployed. E.g., if the identification task is to be utilized in order to promote very specific products to very specific (subgroups), e.g. gender/sex specific products, then one could claim that this task provides merit to an advertising system and does not discriminate. Note though that implicit risks, like gentrification, still exist. On the other hand, if identification is used in a hiring system, then there might be a higher risk of leading to discrimination

%In general, the principles of non-discrimination and AI fairness should be followed. However, these concepts are still defined in very high level and they can be interpreted into a plethora of different, often conflicting fairness definition. Again, in ensuring fairness of AI systems there is no one-size-fits-all solution. Different definitions might be appropriate in different use case scenarios, and in relation with specific policies or interventions that are applicable in such scenarios. "Localization" comprises a form of "personalization": an AI model is able to learn the specificities and patterns of specific subsets of the datasets defined by locality, if proper training features are defined or if separate, locality-based models are learned. Given that, if in specific use cases there exists some kind of merit in learning and deploying locality-specific AI models, then it should be allowed; however, the principles of transparency and explainability should also be considered, as in the personalization case.